%% file: ijcai23.tex
\documentclass{article}
\usepackage{ijcai23}

\usepackage{times}
\usepackage{soul}
\usepackage{url}
\usepackage[hidelinks]{hyperref}
\usepackage[utf8]{inputenc}
\usepackage[small]{caption}
\usepackage{graphicx}
\usepackage{amsmath}
\usepackage{amsthm}
\usepackage{booktabs}
\usepackage{algorithm}
\usepackage[switch]{lineno}

\usepackage{multirow}
\usepackage{microtype}
\usepackage{booktabs}
\usepackage{enumitem}
\usepackage[utf8]{inputenc}
\usepackage{listings}
\usepackage{caption}
\usepackage{dsfont}
\usepackage{algpseudocode}
\usepackage{graphicx}
\usepackage{algorithm}
\usepackage{subfig}
\usepackage{amsthm}
\usepackage{afterpage}
\usepackage{xspace}
\usepackage{adjustbox}

\usepackage{ragged2e} 
\usepackage{booktabs, tabularx}
\usepackage{makecell}
\usepackage{nicematrix,tikz}
\usepackage{wrapfig}

\usepackage{threeparttable}

\newtheorem{objective}{Objective}

\input{math_commands.tex}

\title{Knowledge Distillation on Graphs: A Survey}

\author{
Yijun Tian$^{1,*}$\and
Shichao Pei$^{1,*}$\and
Xiangliang Zhang$^{1}$\and
Chuxu Zhang$^{2}$\and
Nitesh V. Chawla$^1$
\affiliations
$^1$Department of Computer Science, University of Notre Dame, USA\\
$^2$Department of Computer Science, Brandeis University, USA\\
\emails
\{yijun.tian, spei2, xzhang33\}@nd.edu,
chuxuzhang@brandeis.edu,
nchawla@nd.edu
}

\begin{document}
\maketitle

\begin{NoHyper}
\def\thefootnote{*}\footnotetext{equal contribution.}
\end{NoHyper}

\begin{abstract}

Graph Neural Networks (GNNs) have attracted tremendous attention by demonstrating their capability to handle graph data. However, they are difficult to be deployed in resource-limited devices due to model sizes and scalability constraints imposed by the multi-hop data dependency. In addition, real-world graphs usually possess complex structural information and features. Therefore, to improve the applicability of GNNs and fully encode the complicated topological information, knowledge distillation on graphs (KDG) has been introduced to build a smaller yet effective model and exploit more knowledge from data, leading to model compression and performance improvement. Recently, KDG has achieved considerable progress with many studies proposed. In this survey, we systematically review these works. Specifically, we first introduce KDG challenges and bases, then categorize and summarize existing works of KDG by answering the following three questions: 1) what to distillate, 2) who to whom, and 3) how to distillate. Finally, we share our thoughts on future research directions.

\end{abstract}

\section{Introduction}

Graph-structured data is ubiquitous in the real world, with its capability to model a variety of structured and relational systems, such as academic graphs \cite{ogb_datasets}, knowledge graphs \cite{kg_survey_1}, and social networks \cite{social_graph}. To understand and exploit the inherent structure of graphs, Graph Neural Networks (GNNs) have been proposed \cite{gin,gat,gcn}. GNNs have shown exceptional capacity and promising performance in handling non-Euclidean structural data, and have been applied in many downstream applications across different domains, including recommender systems \cite{gnn_rec_1,reciperec}, chemical science \cite{gnn_molecular_survey,gnn_molecular_1}, and food/nutrition services \cite{recipe_rec_frontiers,recipe2vec}.

The success of modern GNNs relies on the complex model structure and the usage of message passing architecture, which aggregates and learns node representations based on their (multi-hop) neighborhood \cite{gnn_survey_1}. However, the advancement of larger and deeper model structures makes training expensive \cite{deepgcns}, and message passing is time-consuming and computation-intensive \cite{glnn}. This poses challenges in applying GNNs to resource-limited devices due to the model size and the scalability constraint. In addition, real-world graphs usually possess complex structures and node features \cite{jin2020graph}, making the encoding of informative topological information a crucial aspect of model designs. Correspondingly, calls for developing new paradigms of resource-friendly and effective GNNs were raised. Fortunately, knowledge distillation (KD) \cite{hinton_kd} has emerged to alleviate the resource limitation by training a smaller model and enhancing the model's capability in encoding data for images \cite{kd_cv}, languages \cite{kd_nlp}, and robotics \cite{kd_robotics}.

\begin{figure}[t]
	\centering
	\includegraphics[width=0.93\columnwidth]{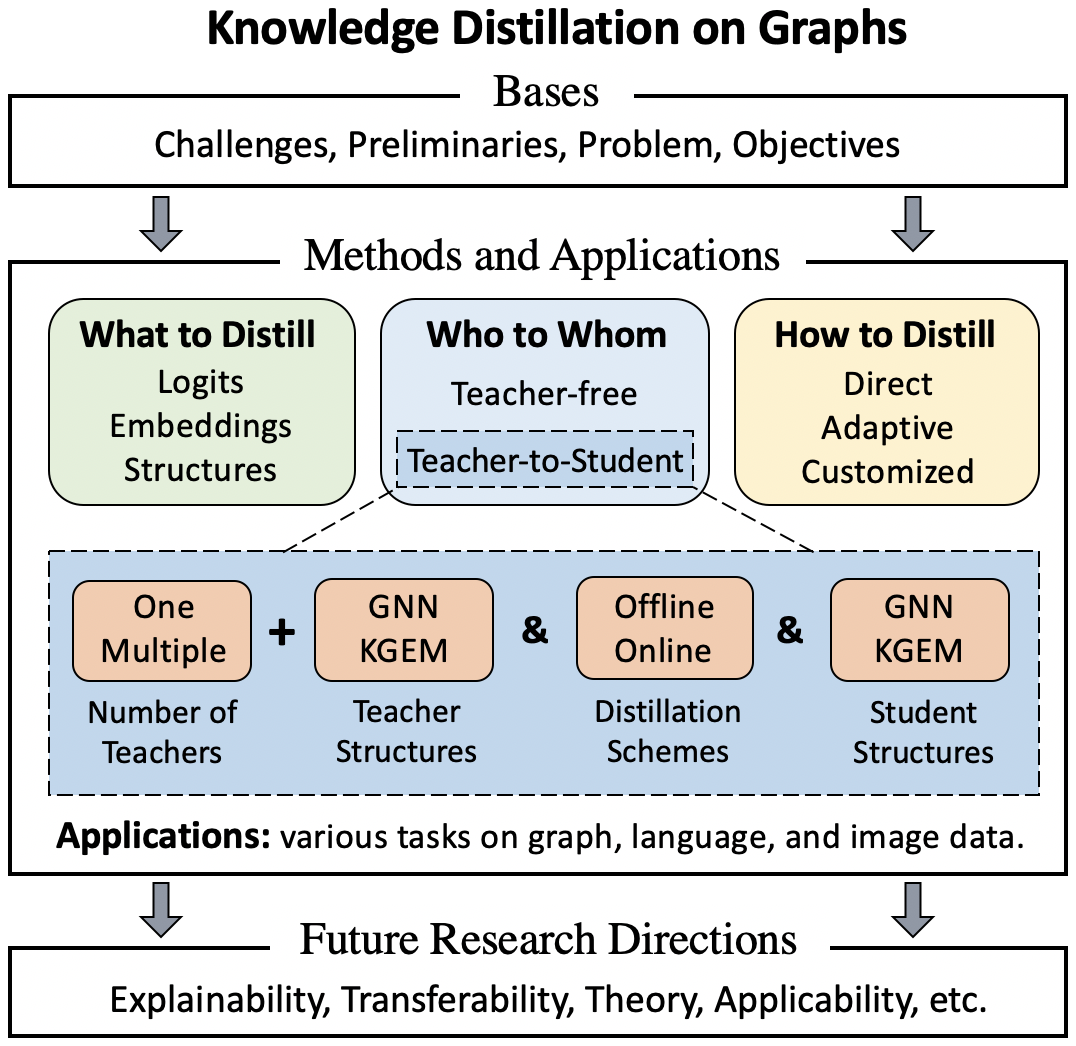}
	\caption{
	The framework of this survey.
	}
	\vspace{-0.1in}
	\label{fig:framework}
\end{figure}

Therefore, knowledge distillation on graphs (KDG), which naturally introduces the advantages of KD into graph learning and builds efficient GNNs, has become a promising research topic and attracted increasing attention from the Artificial Intelligence (AI) community. Consequently, there have been many KDG studies with diverse methodologies and applications in recent years. In this survey, we provide a comprehensive and systematic review of works in KDG, as shown in Figure \ref{fig:framework}. In particular, we start by analyzing the key challenges of KDG. Then we introduce the preliminaries of KDG including graph neural networks and knowledge distillation. After that, we present the formal definition of the KDG problem and describe two common objectives of KDG. Next, we summarize the existing KDG methods by categorizing them with distilled knowledge (e.g., Logits, Embeddings, Structures), distillation directions (e.g., teacher-free, teacher-to-student), and distillation algorithms (e.g, direct, adaptive, customized). We also discuss the impact of three fundamental factors including the number of teachers, the model structures, and the distillation schemes. All in all, we answer the following three questions for KDG: 1) what to distillate, 2) who to whom, and 3) how to distillate. For better comparison and demonstration, we provide a comprehensive summary table listing the representative KDG works with their open-source codes. Finally, we conclude with a discussion of open problems and pressing issues as future research directions of KDG. To summarize, the main contributions of this work are as follows:
\begin{itemize}[leftmargin=*]
    \item This is the first survey paper of KDG, which covers representative KDG methods from 26 publications since 2020.

    \item We systematically survey and categorize existing works by answering three questions: 1) what to distillate, 2) who to whom, and 3) how to distillate.

    \item We discuss the promising future research directions of KDG and encourage further investigations in this field.

\end{itemize}

\section{KDG Challenges and Bases}

KDG, as a new and promising research topic, is non-trivial and faces the following key challenges:

\begin{itemize}[leftmargin=*]
    \item \textbf{The uniqueness of the KDG problem.}
    Unlike general graph learning which focuses on fitting one trainable model to the graph data, KDG aims to train a scalable student model to mimic the teacher model. In addition, KDG differs from KD on images or text since graphs lie in non-Euclidean space with complex topological information. Therefore, novel methods should be designed to solve the problem by considering model scalability and graph data complexity.

    \item \textbf{The complexity of methodology.}
    The success of a KDG method heavily depends on the quality of distilled knowledge, the choice of model structures, the type of distillation schemes, and the appropriate design of distillation algorithms. Therefore, it is essential yet difficult to develop KDG methods with all these components taken into account and achieve the topmost performance by all means.

    \item \textbf{The diversity of downstream applications.}
    KDG methods can be applied to diverse tasks and applications such as natural language inference, image segmentation, and various graph learning tasks (e.g., node and graph classification). Given that different applications may require distinct settings, objectives, constraints, and domain knowledge, it is not simple to develop a customized KDG method tailored to the target application.

\end{itemize}

\noindent
Next, we introduce KDG preliminaries including graph neural networks and knowledge distillation, formally define the problem, and discuss two objectives of KDG.

\subsection{Graph Neural Networks}

Many graph neural networks \cite{gat,gcn,gin} have been proposed to encode the graph-structure data. They utilize the message passing paradigm to learn node embeddings by aggregating information from neighbors. Since GNNs have proven to be exceptionally effective at handling the complexity of structural data, they have become the dominant model backbone for KDG methods. In particular, we define a graph $G$ as $G = (\gV, \gE, X)$, where $\gV$ is the set of nodes, $\gE$ represents the edge set, and $X$ denotes the node features. GNNs learn node embeddings as follows: 
\begin{equation}
    h_v^{l+1} = \text{COM}\left (h_v^{l}, \left[\text{AGG}\left(\left\{h_{u}^{l}~|~\forall u \in \mathcal{N}_v\right\}\right)\right]\right), 
\end{equation}
where $h_v^{l}$, $h_u^{l}$ denotes the embeddings of node $v, u \in \gV$ at $l$-th layer, respectively. $h_v^{l+1}$ is the embedding of node $v$ at $(l+1)$-th layer. $\mathcal{N}_v$ represents the set of neighbors for node $v$. $\text{AGG}(\cdot)$ is the neighbor aggregation function and $\text{COM}(\cdot)$ is the combination function. $h_v^{0}$ is initialized with node attribute $\gX_{v}$. Furthermore, the embedding of the whole graph $\gG$ can be computed as follows:
\begin{equation}
    h_\gG^{l} = \text{READOUT}\left\{h_{v}^{l}~|~\forall v \in \gV\right\}, 
\end{equation}
where the \text{READOUT} function can be a simple permutation invariant function such as summation. Moreover, knowledge graphs, as variants of graphs, have attracted wide attention by regarding entities as nodes and relations between entities as edges \cite{kg_survey_1}. Besides employing GNNs to learn node and edge embeddings, knowledge graph embedding methods \cite{kg_survey_2} are proposed to encode the graph as a collection of fact triplets and learn embeddings via modeling the plausibility score of each triplet.

\subsection{Knowledge Distillation}
Knowledge distillation aims to train a smaller student model by transferring the knowledge from a larger teacher model \cite{kd_survey_1,kd_survey_2}. The main idea is to enforce the student model to mimic the teacher model, where the logits, activations, neurons, and features can all be regarded as the knowledge that guides the learning of the student model \cite{hinton_kd,ahn2019variational,heo2019knowledge}. In particular, the knowledge distillation loss $\mathcal{L}_{kd}$ between the student and the teacher can be defined as follows:
\begin{equation}
    \mathcal{L}_{kd} = \text{DIV}(k^T, k^S), 
    \label{eq:L_kd}
\end{equation}
where \text{DIV} indicates the divergence loss (e.g., Kullback-Leibler divergence), $k^T$ and $k^S$ are the knowledge obtained from the teacher and student models, respectively. A vanilla knowledge distillation considers the logits as the knowledge and employs a Softmax function with temperature to derive the soft targets $p$, which show the probabilities of the input belonging to various classes. Hence, the Eq. \ref{eq:L_kd} can be reformulated as $\mathcal{L}_{kd} = \text{DIV}(p^T, p^S)$, where $p^T$ and $p^S$ represent the soft targets derived from the teacher and students, respectively. Similarly, in the cases where knowledge is not logits, $k^T$ and $k^S$ in Eq. \ref{eq:L_kd} can be replaced accordingly, e.g., $f^T$ and $f^S$ for the learned features of the teacher and student models. After calculating $\mathcal{L}_{kd}$, the student model is trained by a joint objective of both the origin downstream task loss and the knowledge distillation loss $\mathcal{L}_{kd}$, where the former facilitates the student to learn from the original specific task, and the latter targets at transferring the knowledge from the teacher to the student.

\subsection{KDG Problem}

A KDG problem usually holds at least one of the objectives (i.e., model compression, performance improvement), and can be defined as the derivation and combination of appropriate distilled knowledge $K$, model structure $M$, distillation scheme $\mathcal{S}$, and distillation algorithm $\mathcal{A}$ to achieve the objectives. In particular, given a graph $\gG$, a teacher model $M^T$ is presented to take $\gG$ as input to obtain the teacher knowledge $K^T$. Then, a student model $M^S$ is introduced to generate the student knowledge $K^S$ and learn from $K^T$ by comparing the divergence differences between them. The learning procedure is guided by the distillation algorithm $\mathcal{A}$, and the training process (e.g., parameter updating) is defined by $\mathcal{S}$ which can be divided into offline and online distillations. Next, we describe the detailed settings of the two objectives and their differences.

\begin{objective}
{\bf Model Compression.} 
The student model $M^S$ is considered smaller than the teacher model $M^T$, with either fewer intermediate laters, fewer neurons/hidden dimensions in each layer, fewer parameters, or a different model structure with better scalability, e.g., using a multilayer perceptron (MLP) as the student model. After distillation, the student model is more efficiently applicable while still maintaining comparable performances to the teacher model.

\end{objective}

\begin{objective}
{\bf Performance Improvement.} 
The student model $M^S$ has either a smaller, the same, or a different structure as the teacher model $M^T$. The goal is to transfer the pre-acquired knowledge from the teacher to the student, rather than minimize the size of the student. The knowledge obtained from the teacher can benefit the student in capturing the complexities of the data and improving performance compared to the vanilla setting without knowledge distillation.

\end{objective}

\begin{table*}[!ht]
\centering
\resizebox{\textwidth}{!}{
\begin{NiceTabular} {c|c|c|c|c|c|c}
\toprule
\textbf{Method} & \textbf{Objective} & \textbf{What to Distillate} & \textbf{Who to Whom} & \textbf{How to Distillate} & \textbf{Venue} & \textbf{Code Link} \\

 
\midrule
TinyGNN$^{[1]}$ & Compression & Logits & GNN $\Rightarrow$ GNN & Direct & KDD'20 & N/A \\

\midrule
LSP$^{[2]}$ & Compression & Structures & GNN $\Rightarrow$ GNN & Direct & CVPR'20 & \url{shorturl.at/ahQVZ} \\
 
\midrule
RDD$^{[3]}$ & Performance & Logits, Embs & Teacher-free & Adaptive & SIGMOD'20 & N/A \\

\midrule
ROD$^{[4]}$ & Performance & Logits; Structures & Multi. GNNs $\Leftarrow\Rightarrow$ GNN & Adaptive & KDD'21 & \url{shorturl.at/bfkor} \\

\midrule
GNN-SD$^{[5]}$ & Performance & Structures & Teacher-free & Adaptive & IJCAI'21 & N/A \\

\midrule
CPF$^{[6]}$ & Compression & Logits & GNN $\Rightarrow$ MLP & Direct & WWW'21 & \url{shorturl.at/hLWX6} \\

\midrule
EDGE$^{[7]}$ & Performance & Structures, Embs & GNN $\Rightarrow$ GNN & Direct & NAACL'21 & N/A \\
  
\midrule
GFKD$^{[8]}$ & Compression & Logits & GNN $\Rightarrow$ GNN & Direct & IJCAI'21 & \url{shorturl.at/nEGOX} \\

\midrule
MulDE$^{[9]}$ & Compression & Logits & Multi. KGEMs $\Rightarrow$ KGEM & Adaptive & WWW'21 & N/A \\

\midrule
GLNN$^{[10]}$ & Compression & Logits & GNN $\Rightarrow$ MLP & Direct & ICLR'22 & \url{shorturl.at/aCHLY} \\
 
\midrule
MSKD$^{[11]}$ & Compression & Logits; Structures & Multi. GNNs $\Rightarrow$ GNN & Adaptive & AAAI'22 & \url{shorturl.at/fpwD4} \\
 
\midrule
GraphAKD$^{[12]}$ & Compression & Logits; Embs & GNN $\Rightarrow$ GNN & Customized & KDD'22 & \url{shorturl.at/cdjPS} \\
  
\midrule
FreeKD$^{[13]}$ & Performance & Logits; Structures & GNN $\Leftarrow\Rightarrow$ GNN & Adaptive & KDD'22 & N/A \\
  
\midrule
CKD$^{[14]}$ & Performance & Structures & Teacher-free & Customized & WWW'22 & \url{shorturl.at/kuCM2} \\
  
\midrule
G-CRD$^{[15]}$ & Performance & Structures & GNN $\Rightarrow$ GNN & Customized & TNNLS'22 & N/A \\

\midrule
Cold Brew$^{[16]}$ & Compression & Structures & GNN $\Rightarrow$ MLP & Direct & ICLR'22 & \url{shorturl.at/fmGN3} \\

\midrule
DFAD-GNN$^{[17]}$ & Compression & Logits & GNN $\Rightarrow$ GNN & Customized & IJCAI'22 & N/A \\

\midrule
Alignahead$^{[18]}$ & Performance & Structures & GNN $\Leftarrow\Rightarrow$ GNN & Direct & IJCNN'22 & \url{shorturl.at/imzH4} \\
   
\midrule
KDGA$^{[19]}$ & Performance & Logits & GNN $\Rightarrow$ GNN & Direct & NeurIPS'22 & \url{shorturl.at/gvKNP} \\
    
\midrule
GKD$^{[20]}$ & Compression & Structures & GNN $\Rightarrow$ GNN & Direct & NeurIPS'22 & \url{shorturl.at/kOXZ5} \\

\midrule
SAIL$^{[21]}$ & Performance & Embs & GNN $\Rightarrow$ GNN & Direct & AAAI'22 & N/A \\

\midrule
LTE4G$^{[22]}$ & Performance & Logits & Multi. GNNs $\Rightarrow$ GNN & Direct & CIKM'22 & \url{shorturl.at/ilpzM} \\


\midrule
T2-GNN$^{[23]}$ & Performance & Logits; Embs & Multi. GNNs $\Rightarrow$ GNN & Direct & AAAI'23 & N/A \\
    
\midrule
RELIANT$^{[24]}$ & Compression & Logits & GNN $\Rightarrow$ GNN & Direct & SDM'23 & \url{shorturl.at/fJL17} \\
   
\midrule
BGNN$^{[25]}$ & Performance & Logits & Multi. GNNs $\Rightarrow$ GNN & Adaptive & AAAI'23 & N/A \\

\midrule
NOSMOG$^{[26]}$ & Compression & Logits; Structures & GNN $\Rightarrow$ MLP & Direct & ICLR'23 & \url{shorturl.at/ceFJS} \\

\bottomrule
\end{NiceTabular}
}
\begin{tablenotes}[flushleft]
\item 
\small{Note: }
\scriptsize{
$^{[1]}$\cite{tinygnn}; $^{[2]}$\cite{lsp};  $^{[3]}$\cite{rdd}; $^{[4]}$\cite{rod}; $^{[5]}$\cite{gnnsd}; $^{[6]}$\cite{cpf}; $^{[7]}$\cite{edge}; $^{[8]}$\cite{gfkd}; $^{[9]}$\cite{mulde}; $^{[10]}$\cite{glnn}; $^{[11]}$\cite{mskd}; $^{[12]}$\cite{graphakd}; $^{[13]}$\cite{freekd}; $^{[14]}$\cite{ckd}; $^{[15]}$\cite{gcrd}; $^{[16]}$\cite{cold_brew}; $^{[17]}$\cite{dfadgnn}; $^{[18]}$\cite{alignahead}; $^{[19]}$\cite{kdga}; $^{[20]}$\cite{gkd}; $^{[21]}$\cite{sail}; $^{[22]}$\cite{lte4g}; $^{[23]}$\cite{t2gnn}; $^{[24]}$\cite{reliant}; $^{[25]}$\cite{bgnn}; $^{[26]}$\cite{nosmog}.
}
\end{tablenotes}
\caption{
A list of representative KDG methods with open-source code. `Who to Whom' holds the format: (teacher)(offline distillation $\Rightarrow$, or online distillation $\Leftarrow\Rightarrow$)(student). Embs stands for embeddings and KGEM represents knowledge graph embedding method.
} 
\label{tab:summary}
\end{table*}

\section{Methods and Applications}
In this section, we thoroughly review the existing works of KDG (Table \ref{tab:summary}) and answer the following three questions: 1) what to distillate, 2) who to whom, and 3) how to distillate. Then we examine and introduce the relevant applications.

\subsection{What to Distillate}
The primary step of KDG is to decide what knowledge to distill. There are three kinds of information that can be regarded as transferable knowledge for distillation on graphs, i.e., output logits, graph structures, and embeddings. Next, we introduce each of the knowledge sources with related works.

\noindent
\textbf{Logits.}
The logits denote the inputs to the final Softmax and represent the soft label prediction. For downstream tasks, the probability distributions/scores can be obtained by feeding the logits to a Softmax layer \cite{tinygnn,mulde}. After calculating the logits, the KDG methods minimize the difference between the probability distribution (or scores) of a teacher model and a student model to distill the knowledge. There are various ways to measure the differences. For example, a number of works adopt KL diversity \cite{rod,mulde,gfkd,glnn,mskd,freekd,kdga,lte4g,t2gnn,bgnn,nosmog}, while several other works employ soft cross-entropy loss \cite{tinygnn,graphakd} and statistical distances \cite{cpf,dfadgnn,reliant}. Conducting the knowledge distillation via logits enables the student model to learn from the teacher from the output space, which also provides additional soft labels as training targets, especially in the semi-supervised learning setting where the label supervisions can be produced to a large number of unlabeled data samples.

\noindent
\textbf{Structures.}
Graph structure depicts the connectivity and relationships between the elements of a graph, such as nodes and edges, and plays a crucial role in graph data modeling. Therefore, how to preserve and distill graph structure information from a teacher model to a student model has been a focus in recent studies. In particular, several works \cite{lsp,edge,freekd,alignahead} design strategies to formulate the local structures of nodes and distill them into the student model, so that the student can learn the knowledge of how the teacher model describes the relationship between a node and its neighbors. For example, LSP \cite{lsp} first presents to model the local structures via vectors that calculate the similarity between nodes and their one-hop neighbors, where later FreeKD \cite{freekd} and Alignahead \cite{alignahead} follow this design to distill the structural information into the student. Besides, GNN-SD \cite{gnnsd} represents neighborhood discrepancy as local structures and focuses on distilling them from shallow GNN layers to deep ones in order to preserve the non-smoothness of the embedded graph on deep GNNs.

However, the local structures only provide limited local structural information around specific nodes and are incapable of representing the whole picture of a graph. Recent works \cite{rod,mskd,ckd,nosmog} attempt to overcome this shortcoming and focus on capturing global structures to complement the local structures. For instance, CKD \cite{ckd} works on heterogeneous graphs and proposes to model the global structure information via meta-paths. G-CRD \cite{gcrd} leverages contrastive learning to implicitly preserve global topology by aligning the student node embeddings to those of the teacher in a shared representation space. ROD \cite{rod} proposes to encode knowledge on different-scale reception fields by encoding graphs at various levels of localities using multiple student models, so that a more powerful teacher can be assembled from these students to capture multi-scale knowledge and provide rich information to the student. Similarly, MSKD \cite{mskd} tries to capture the topological semantics at different scales to supervise the training of the student model. Furthermore, Cold Brew \cite{cold_brew}, GKD \cite{gkd}, and NOSMOG \cite{nosmog} propose to use structural embedding, neural heat kernel, and positional encoding to capture the global structure, respectively. The distillation of graph structural information enables the student model to retain the knowledge of topological connections and node relationships from the teacher, thereby facilitating a comprehensive encoding of the complex structure and further enhancing the performance.

\noindent
\textbf{Embeddings.}
Instead of logits and graph structures, several efforts adopt the learned node embeddings from the intermediate layers of teacher models to guide the learning of the student model. To illustrate, T2-GNN and SAIL \cite{t2gnn,sail} maximize the consistency of embeddings of the same node from the teacher model and the student model. SAIL \cite{sail} aims to push the node embeddings and node features to be consistent. GraphAKD \cite{graphakd} leverages adversarial learning to distill node representations of the teacher model from both local and global views to the student model. RDD \cite{rdd} enforces the student model to directly mimic the entire node embeddings of the teacher model for more informative knowledge than merely using the Softmax outputs. The usage of embeddings for distillation enables the student to observe the learning process of the teacher through the derivation and transformation of the embeddings. It provides an alternative perspective to guide the student in learning representation, other than imitating the labels in output space or relational topological information.

\noindent
\textbf{Comparison.}
It is essential to understand the impact of each type of knowledge and how different kinds of knowledge benefit the student model in a complementary manner. To demonstrate, using logits as distilled knowledge has a similar motivation and effect to label smoothing and model regularization \cite{label_smoothing,ding2019adaptive}, which also offers additional training targets in addition to the ground truth labels. Structured-based knowledge exploits the relationships between nodes to capture the complex graph topological information, forcing the student model to focus more on graph structures than information in the output space and hidden layers. Embeddings as distilled knowledge enable the student model to mimic the intermediate workflow of the teacher model, providing extra knowledge of the learning process that happened prior to the generation of the final output.

\subsection{Who to Whom}

After determining what knowledge to distill, we need to define the teacher and student as well as decide the distillation schemes. In particular, we discuss them from the perspectives of teacher-free and teacher-to-student.

\noindent
\textbf{Teacher-free.}
Teacher-free corresponds to the situation that a student model can learn knowledge from itself without an external teacher model. Usually, teacher-free knowledge distillation refers to self-distillation based on the knowledge extracted from a single model such as distilling knowledge between different layers \cite{gnnsd} or different graph structures \cite{ckd}. Specifically, with the aim of overcoming the over-smoothing issue, GNN-SD \cite{gnnsd} extracts and transfers the neighborhood discrepancy knowledge between layers of a single network in order to distinguish node embeddings in deep layers from their neighbor embeddings. CKD \cite{ckd} develops two collaborative distillations including an intra-meta-path module to distill the regional and global knowledge within each meta-path and an inter-meta-path module to transfer the regional and global patterns among different meta-paths. In addition, some methods build an ensemble teacher using multiple versions of the model itself without introducing an external teacher model while still benefiting from the advantages of distillation. For example, RDD \cite{rdd} presents a self-boosting framework based on ensemble learning to incorporate the node and edge reliabilities. Teacher-free alleviates the need for an external teacher model and enforces the model to focus on the in-depth knowledge utilization in the model itself.

\noindent
\textbf{Teacher-to-Student.}
The teacher-to-student distillation framework aims to distill knowledge from one or multiple teacher models to a student model. To demonstrate, we analyze and compare the impact of teacher numbers, different model structures, and offline and online distillations.

\noindent
\textit{\underline{One or multiple teachers.}}
Most KDG methods rely on the standard one-teacher framework to conduct knowledge distillation \cite{tinygnn,lsp,glnn,gkd,sail,nosmog}, in which they try to transfer information from a well-trained teacher to the student. For example, NOSMOG \cite{nosmog} intends to distill an MLP student from a pre-trained GNN teacher. Recently, several works investigate the use of multiple teacher models to provide comprehensive knowledge for the student model. For example, ROD \cite{rod} and MSKD \cite{mskd} leverage multiple teachers to encode the different levels of localities or different topological semantics at different scales. MulDE \cite{mulde} introduces a group of hyperbolic KGE models and integrates their knowledge to train a smaller KGE model. LTE4G \cite{lte4g} assigns multiple expert GNN models to different subsets of nodes considering both the class and degree long-tailedness. T2-GNN \cite{t2gnn} proposes two teacher models to provide feature-level and structure-level guidances for the student. BGNN \cite{bgnn} takes the multiple generated GNN models from previous steps as teachers to supervise the training of the student model. Multiple teachers may transfer more knowledge than a single teacher, but they will likely introduce noisy knowledge into the student model, thereby decreasing its robustness.

\noindent
\textit{\underline{Teacher and student structures.}}
KDG approaches adopt different teacher and student structures for distinct purposes. First, some KDG methods distill the knowledge from a larger GNN teacher to a smaller GNN student for compression \cite{tinygnn,lsp,graphakd,gcrd,dfadgnn}. Second, with the target of lessening the dependence on the graph structure and reducing GNN latency caused by message passing, several works introduce an MLP as the student and achieve outstanding performance \cite{nosmog,glnn,cold_brew}. Third, a number of KDG models focus on improving the model performance in situations where the student holds the same structure as the teacher. To demonstrate, FreeKD \cite{freekd} and Alignahead \cite{alignahead} train two shallower GNNs simultaneously to guide each other with each one can be the teacher or student. KDGA \cite{kdga} and EDGE \cite{edge} adopt the same structures for the teacher and student but with different input data. In particular, KDGA utilizes the node features and augmented structure as input to the teacher, whereas the student only takes the original structure. EDGE takes an augmented knowledge graph with text information as input to the teacher while the student model uses the original knowledge graph without additional information. Despite having the same structure, these methods resort to richer data to provide supplementary knowledge to improve the student. Furthermore, KGEM methods are employed to fully capture the triplets connections in knowledge graphs \cite{mulde}.

\noindent
\textit{\underline{Offline and online distillations.}}
Online distillation refers to the distillation scheme in which both the teacher model and the student model are trained end-to-end, as opposed to offline distillation, in which the teacher model is pre-trained and used to facilitate the training of students without any additional update for the teacher. The majority of existing KDG methods adhere to the prevalent offline distillation \cite{glnn,lsp,tinygnn,kdga}, whereas only a small number of studies attempt to design the online distillation scheme. For example, ROD \cite{rod} proposes to assemble a teacher model using predictions of multiple students and leverages the ensemble teacher to train these students. In this way, the teacher and students can be updated alternatively. FreeKD \cite{freekd} builds two collaboratively shallower GNNs with the desire to exchange knowledge between them via reinforcement learning, so that they can be optimized together and enhance each other. Similarly, Alignahead \cite{alignahead} devises two student models and updates them in an alternating training procedure, in order to circumvent the demand for the pre-trained teacher model. In general, the teacher-to-student framework has been the basis of most KDG methods, which can be easily tailored to achieve the two objectives and extended for various applications.

\noindent
\textbf{Comparison.}
Compared to condensing GNNs, the teacher-free structure enables the student to distill knowledge by itself to achieve better performance, thereby avoiding the dependency on another teacher model and sidestepping the heavy computation costs. The reasons for better performance achievement are that distilling knowledge from the student itself can help the student converge to flat minima and prevent the vanishing gradient problem \cite{zhang2019your}. On the other hand, the teacher-to-student structure naturally fits the objective of model compression where a smaller student can be obtained, while still maintaining the capability to improve performance. For the number of teachers, employing multiple teachers enables the student to fuse information from diverse sources to establish a comprehensive understanding of the knowledge, similar to ensemble learning. However, how to trade off the instruction from different teachers and integrate them without performance degradation needs special designs. In contrast, a single teacher is sometimes better and more convenient by bypassing the intricacy of incorporating multiple ones. For the determination of teacher and student structures for KDG methods, GNNs demonstrate outstanding capability in capturing graph topological information and have become the most popular choice. Moreover, some methods introduce MLP as the student model to circumvent the scalability limitation imposed by GNNs or utilize KGEM to adapt to knowledge graphs. As long as the model performs well with the target graph data, the choice of the model structure is typically not a crucial aspect of KDG methods. For distillation schemes, offline distillation is typically used to transfer knowledge from a complex teacher model, whereas teacher and student models are comparable and can be trained concurrently in the setting of online distillation.

\subsection{How to Distillate}
After selecting the type of knowledge and determining the strategy of who to whom \textit{w.r.t.} the teacher and student in the distillation, the next essential question is to decide how to develop an effective algorithm for distilling the obtained knowledge. In general, existing distillation algorithms can be categorized into three types: direct, adaptive, and customized distillations. We introduce them in the following discussion.

\noindent
\textbf{Direct.}
Direct distillation refers to a type of distillation algorithm where the divergences between the knowledge of the teacher and the student are directly minimized to force the student fully mimic the teacher model. For example, many studies directly distill the node logits from the teacher to the student \cite{tinygnn,glnn,kdga,lte4g,t2gnn,reliant,nosmog}, while some others focus on directly distilling the node embeddings \cite{edge,cold_brew,sail}. Compared to node classification in which the node logits are utilized, certain graph classification methods conduct the distillation by comparing the logits of all graphs \cite{gfkd,dfadgnn}. In addition, LSP \cite{lsp} and Alignahead \cite{alignahead} directly minimize the divergence of local structures between the teacher and the student. To preserve the graph structures, GKD \cite{gkd} directly minimizes the Frobenius distance between Neural Heat Kernel matrices on the teacher and the student. Generally, direct distillation can be implemented easily to facilitate the straightforward knowledge transfer and  regularize the training of the student.

\noindent
\textbf{Adaptive.}
Different from direct distillation, adaptive distillation provides a more flexible paradigm for conducting distillation by adaptively considering the significance of knowledge. For example, RDD \cite{rdd} proposes to distill the knowledge that the teacher learns reliably while the student learns incorrectly. FreeKD \cite{freekd} introduces an adaptive strategy to determine the selection of distillation direction as well as the propagated local structures between two shallower GNNs. GNN-SD \cite{gnnsd} performs the self-distillation only when the magnitude of neighborhood discrepancy of the target layer is larger than that of the online layer. In addition, some KDG methods involving multiple teacher models employ an adaptive strategy to combine the knowledge of various teachers. For instance, ROD \cite{rod} leverages a gate component to combine knowledge over different levels of localities. MulDE \cite{mulde} adaptively aggregates the prediction from multiple teachers and returns soft labels for students with relation-specific scaling and contrast attention mechanisms. MSKD \cite{mskd} adaptively exploits the local structures from multiple teachers with an attentional topological semantic mapping design. Furthermore, a recent study BGNN \cite{bgnn} adaptively assigns value to the hyper-parameter temperature for each node with learnable parameters. Empowered by adaptive distillation, the informative knowledge can be selected and aggregated to train a more effective student model. Unlike direct distillation, adaptive distillation provides a more flexible and robust paradigm for mitigating the negative impact of less-informative or noisy knowledge.

\noindent
\textbf{Customized.}
Apart from direct distillation and adaptive distillation, there are several methods adopting a variety of machine learning techniques to distill knowledge. For example, GraphAKD \cite{graphakd} and DFAD-GNN \cite{dfadgnn} follow the principle of adversarial learning to distinguish the student and teacher using a trainable discriminator instead of forcing the student network to precisely mimic the teacher network with manually designed distance functions. CKD \cite{ckd} proposes a collaborative KD method to incorporate the knowledge extracted within and between the meta-paths. G-CRD \cite{gcrd} formulates the representation distillation as a contrastive learning task on pairwise relationships across the teacher and student embedding spaces. Typically, customized distillation enables diversified designs of KDG methods according to the distinct objective of tasks in different scenarios, which stimulates further studies to explore more distillation strategies.

\noindent
\textbf{Comparison.}
Although direct distillation can achieve satisfactory performances in most cases by allowing the student model to directly mimic the teacher, it disregards the distinct importance of different knowledge, such as the fact that not all nodes in a graph contain the same amount of information. Therefore, even though direct distillation has the benefit of being simple and straightforward, merely using direct distillation can be sub-optimal and lack flexibility. Compared to direct distillation, adaptive distillation is more flexible by determining the weight and significance of different knowledge and combining them adaptively. This approach avoids the direct distillation's incapability to distinguish between different types of knowledge, but inevitably introduces additional computation costs, such as the dependency on a weighted gating module and the use of special distillation strategies. In addition, customized distillation provides another option for distilling the knowledge including using contrastive learning to implicitly preserve the graph topology, allowing researchers to develop personalized solutions for the target problem.

\subsection{Applications}

As an effective technique in compressing the GNNs and improving their performance, KDG methods have been widely applied in various applications such as graph learning tasks and different fields of AI. For example, KDG methods can be naturally applied to node classification, graph classification, and link prediction tasks \cite{nosmog,cold_brew}. The node classification task aims to predict the category of nodes. Some common public benchmark datasets \cite{glnn} are as follows: Cora, CiteSeer, and PubMed. The graph classification task focuses on predicting the label of graphs. Some example datasets are PROTEINS and Molhiv \cite{ogb_datasets}. The link prediction task attempts to predict whether two given nodes are connected by an edge \cite{fakeedge}, with datasets usually the same as those in the node classification task. In addition, KDG methods can be applied to various tasks on language and image data such as text classification \cite{text_classification_1} and image classification \cite{hkd}. To demonstrate, these works build a graph from the text and image data and then apply KDG methods to capture the relationships between words, sentences, or images. An example is constructing a graph by connecting messages in different languages from the node and semantic levels for multilingual social event detection \cite{social_event_detection}. Moreover, KDG methods can be introduced to jointly model multimodal content and relational information such as detecting illicit drug trafficker on social media \cite{detect_illicit_drug_trafficker_1}.

\section{KDG Future Research Directions}

KDG is an emerging and rapidly developing research topic. Despite the recent success and the significant progress in KDG, there are still many challenges to be solved. This opens up a number of opportunities for future research directions. In this section, we describe and suggest some of them as follows:

\begin{itemize}[leftmargin=*]

\item \textbf{Explainability.}
Existing KDG methods aim at developing a condensed model or improving the model performance while none of them consider the model explainability. However, developing KDG methods with explainability is crucial for enhancing user trust and model dependability. For instance, it is worthwhile to investigate and explain why and how a particular distillation algorithm works well for a target application, so we can have better justifications for model designs. Potential approaches to provide interpretability include  training an explainable student model \cite{egnn_explanation} or building another model to quantitatively explain the knowledge contributions \cite{haselhoff2021towards}.

\item \textbf{Transferability.}
Typically, the student model acquires domain-specific knowledge from the teacher and then leverages the knowledge to address a target problem. Although the knowledge can be beneficial for the target task, it may not apply to other tasks, thereby limiting the transferability of the student model. Therefore, how to determine the transferable common knowledge from the teacher and distill it to the student for various downstream applications require investigation. Designing distillation algorithms with better generalisability \cite{he2022knowledge} or exploring transferable GNNs \cite{transferable_gnn} are potential solutions.

\item \textbf{Theoretical framework.}
Despite a huge number of KDG methods, the understanding of KDG such as theoretical analyses has not been investigated. To provide guidance for future method development and build a solid foundation, a sound theoretical framework is necessary and important for us to better comprehend KDG methods. In particular, a number of recent works on the knowledge distillation theory \cite{kd_theory_1,kd_theory_2} can serve as foundations for this research avenue.

\item \textbf{Applicability.}
Existing KDG methods typically work on homogeneous graphs, while many other types of graphs are less investigated such as heterogeneous graphs \cite{hgmae} and temporal graphs \cite{ma2022hierarchical}. To improve the applicability of KDG methods to various types of graphs, it is essential to take into account different graph properties, such as the edges types and attributes, as well as to explore how to extract qualitative knowledge and implement an efficient distillation algorithm. Hence, utilizing appropriate distillation techniques for graphs with different properties is also a promising research direction.

\item \textbf{Graph Distillation.}
Similar to distilling the knowledge from GNNs, knowledge from the graph data can also be distilled, i.e., graph distillation. The goal is to synthesize a small graph so that GNNs trained on top of it can achieve comparable performance while being extremely efficient. This line of research can benefit from the matching of training trajectories \cite{cazenavette2022dataset}. In addition, techniques of graph pruning \cite{chen2021unified} and graph condensation \cite{graph_condensation} can be compared and introduced to distill comprehensive knowledge as well as construct a graph with rich information.

\end{itemize}

\section{Conclusion}

Knowledge distillation on graphs (KDG), as an emerging research field, has attracted extensive attention and plays an important role in various application domains. This paper presents the first comprehensive survey on KDG. In particular, we introduce the bases including challenges and preliminaries as well as formally define the problem and objectives. We also thoroughly discuss, categorize, and summarize the existing works according to the proposed taxonomy. In addition, we share our thoughts on future directions. We hope this paper serves as a useful resource for researchers and advances the future work of KDG.

\clearpage
\small
\bibliographystyle{named}
\bibliography{reference}

\end{document}

%% file: math_commands.tex
\usepackage{amsmath,amsfonts,bm}

\def\eqref#1{equation~\ref{#1}}

\def\1{\bm{1}}

\DeclareMathAlphabet{\mathsfit}{\encodingdefault}{\sfdefault}{m}{sl}
\SetMathAlphabet{\mathsfit}{bold}{\encodingdefault}{\sfdefault}{bx}{n}

\def\gE{{\mathcal{E}}}

\def\gG{{\mathcal{G}}}

\def\gV{{\mathcal{V}}}

\def\gX{{\mathcal{X}}}












%% file: ijcai23.bbl
\begin{thebibliography}{}

\bibitem[\protect\citeauthoryear{Ahn \bgroup \em et al.\egroup
  }{2019}]{ahn2019variational}
Sungsoo Ahn, Shell~Xu Hu, Andreas Damianou, Neil~D Lawrence, and Zhenwen Dai.
\newblock Variational information distillation for knowledge transfer.
\newblock In {\em CVPR}, 2019.

\bibitem[\protect\citeauthoryear{Allen-Zhu and Li}{2020}]{kd_theory_1}
Zeyuan Allen-Zhu and Yuanzhi Li.
\newblock Towards understanding ensemble, knowledge distillation and
  self-distillation in deep learning.
\newblock In {\em ICLR}, 2020.

\bibitem[\protect\citeauthoryear{Bian \bgroup \em et al.\egroup
  }{2020}]{social_graph}
Tian Bian, Xi~Xiao, Tingyang Xu, Peilin Zhao, Wenbing Huang, Yu~Rong, and
  Junzhou Huang.
\newblock Rumor detection on social media with bi-directional graph
  convolutional networks.
\newblock In {\em AAAI}, 2020.

\bibitem[\protect\citeauthoryear{Cazenavette \bgroup \em et al.\egroup
  }{2022}]{cazenavette2022dataset}
George Cazenavette, Tongzhou Wang, Antonio Torralba, Alexei~A Efros, and
  Jun-Yan Zhu.
\newblock Dataset distillation by matching training trajectories.
\newblock In {\em CVPR}, 2022.

\bibitem[\protect\citeauthoryear{Chen \bgroup \em et al.\egroup
  }{2021a}]{chen2021unified}
Tianlong Chen, Yongduo Sui, Xuxi Chen, Aston Zhang, and Zhangyang Wang.
\newblock A unified lottery ticket hypothesis for graph neural networks.
\newblock In {\em ICML}, 2021.

\bibitem[\protect\citeauthoryear{Chen \bgroup \em et al.\egroup
  }{2021b}]{gnnsd}
Yuzhao Chen, Yatao Bian, Xi~Xiao, Yu~Rong, Tingyang Xu, and Junzhou Huang.
\newblock On self-distilling graph neural network.
\newblock In {\em IJCAI}, 2021.

\bibitem[\protect\citeauthoryear{Deng and Zhang}{2021}]{gfkd}
Xiang Deng and Zhongfei Zhang.
\newblock Graph-free knowledge distillation for graph neural networks.
\newblock In {\em IJCAI}, 2021.

\bibitem[\protect\citeauthoryear{Ding \bgroup \em et al.\egroup
  }{2019}]{ding2019adaptive}
Qianggang Ding, Sifan Wu, Hao Sun, Jiadong Guo, and Shu-Tao Xia.
\newblock Adaptive regularization of labels.
\newblock {\em arXiv preprint arXiv:1908.05474}, 2019.

\bibitem[\protect\citeauthoryear{Dong \bgroup \em et al.\egroup
  }{2022}]{fakeedge}
Kaiwen Dong, Yijun Tian, Zhichun Guo, Yang Yang, and Nitesh Chawla.
\newblock Fakeedge: Alleviate dataset shift in link prediction.
\newblock In {\em LoG}, 2022.

\bibitem[\protect\citeauthoryear{Dong \bgroup \em et al.\egroup
  }{2023}]{reliant}
Yushun Dong, Binchi Zhang, Yiling Yuan, Na~Zou, Qi~Wang, and Jundong Li.
\newblock Reliant: Fair knowledge distillation for graph neural networks.
\newblock {\em SDM}, 2023.

\bibitem[\protect\citeauthoryear{Fan \bgroup \em et al.\egroup
  }{2019}]{gnn_rec_1}
Wenqi Fan, Yao Ma, Qing Li, Yuan He, Eric Zhao, Jiliang Tang, and Dawei Yin.
\newblock Graph neural networks for social recommendation.
\newblock In {\em WWW}, 2019.

\bibitem[\protect\citeauthoryear{Feng \bgroup \em et al.\egroup
  }{2022}]{freekd}
Kaituo Feng, Changsheng Li, Ye~Yuan, and Guoren Wang.
\newblock Freekd: Free-direction knowledge distillation for graph neural
  networks.
\newblock In {\em KDD}, 2022.

\bibitem[\protect\citeauthoryear{Gou \bgroup \em et al.\egroup
  }{2021}]{kd_survey_1}
Jianping Gou, Baosheng Yu, Stephen~J Maybank, and Dacheng Tao.
\newblock Knowledge distillation: A survey.
\newblock {\em International Journal of Computer Vision}, 2021.

\bibitem[\protect\citeauthoryear{Guo \bgroup \em et al.\egroup
  }{2022a}]{alignahead}
Jiongyu Guo, Defang Chen, and Can Wang.
\newblock Alignahead: Online cross-layer knowledge extraction on graph neural
  networks.
\newblock In {\em IJCNN}, 2022.

\bibitem[\protect\citeauthoryear{Guo \bgroup \em et al.\egroup
  }{2022b}]{gnn_molecular_survey}
Zhichun Guo, Bozhao Nan, Yijun Tian, Olaf Wiest, Chuxu Zhang, and Nitesh~V
  Chawla.
\newblock Graph-based molecular representation learning.
\newblock {\em arXiv preprint arXiv:2207.04869}, 2022.

\bibitem[\protect\citeauthoryear{Guo \bgroup \em et al.\egroup }{2023}]{bgnn}
Zhichun Guo, Chunhui Zhang, Yujie Fan, Yijun Tian, Chuxu Zhang, and Nitesh
  Chawla.
\newblock Boosting graph neural networks via adaptive knowledge distillation.
\newblock In {\em AAAI}, 2023.

\bibitem[\protect\citeauthoryear{Haselhoff \bgroup \em et al.\egroup
  }{2021}]{haselhoff2021towards}
Anselm Haselhoff, Jan Kronenberger, Fabian Kuppers, and Jonas Schneider.
\newblock Towards black-box explainability with gaussian discriminant knowledge
  distillation.
\newblock In {\em CVPR}, 2021.

\bibitem[\protect\citeauthoryear{He \bgroup \em et al.\egroup
  }{2022a}]{graphakd}
Huarui He, Jie Wang, Zhanqiu Zhang, and Feng Wu.
\newblock Compressing deep graph neural networks via adversarial knowledge
  distillation.
\newblock In {\em KDD}, 2022.

\bibitem[\protect\citeauthoryear{He \bgroup \em et al.\egroup
  }{2022b}]{he2022knowledge}
Ruifei He, Shuyang Sun, Jihan Yang, Song Bai, and Xiaojuan Qi.
\newblock Knowledge distillation as efficient pre-training: Faster convergence,
  higher data-efficiency, and better transferability.
\newblock In {\em CVPR}, 2022.

\bibitem[\protect\citeauthoryear{Heo \bgroup \em et al.\egroup
  }{2019}]{heo2019knowledge}
Byeongho Heo, Minsik Lee, Sangdoo Yun, and Jin~Young Choi.
\newblock Knowledge transfer via distillation of activation boundaries formed
  by hidden neurons.
\newblock In {\em AAAI}, 2019.

\bibitem[\protect\citeauthoryear{Hinton \bgroup \em et al.\egroup
  }{2015}]{hinton_kd}
Geoffrey Hinton, Oriol Vinyals, Jeff Dean, et~al.
\newblock Distilling the knowledge in a neural network.
\newblock {\em arXiv preprint arXiv:1503.02531}, 2015.

\bibitem[\protect\citeauthoryear{Hogan \bgroup \em et al.\egroup
  }{2021}]{kg_survey_1}
Aidan Hogan, Eva Blomqvist, Michael Cochez, Claudia d’Amato, Gerard~de Melo,
  Claudio Gutierrez, Sabrina Kirrane, Jose Emilio~Labra Gayo, Roberto Navigli,
  Sebastian Neumaier, et~al.
\newblock Knowledge graphs.
\newblock {\em ACM Computing Surveys}, 2021.

\bibitem[\protect\citeauthoryear{Hu \bgroup \em et al.\egroup
  }{2020}]{ogb_datasets}
Weihua Hu, Matthias Fey, Marinka Zitnik, Yuxiao Dong, Hongyu Ren, Bowen Liu,
  Michele Catasta, and Jure Leskovec.
\newblock Open graph benchmark: Datasets for machine learning on graphs.
\newblock In {\em NeurIPS}, 2020.

\bibitem[\protect\citeauthoryear{Huo \bgroup \em et al.\egroup }{2023}]{t2gnn}
Cuiying Huo, Di~Jin, Yawen Li, Dongxiao He, Yu-Bin Yang, and Lingfei Wu.
\newblock T2-gnn: Graph neural networks for graphs with incomplete features and
  structure via teacher-student distillation.
\newblock In {\em AAAI}, 2023.

\bibitem[\protect\citeauthoryear{Jin \bgroup \em et al.\egroup
  }{2020}]{jin2020graph}
Wei Jin, Yao Ma, Xiaorui Liu, Xianfeng Tang, Suhang Wang, and Jiliang Tang.
\newblock Graph structure learning for robust graph neural networks.
\newblock In {\em KDD}, 2020.

\bibitem[\protect\citeauthoryear{Jin \bgroup \em et al.\egroup
  }{2021}]{graph_condensation}
Wei Jin, Lingxiao Zhao, Shichang Zhang, Yozen Liu, Jiliang Tang, and Neil Shah.
\newblock Graph condensation for graph neural networks.
\newblock In {\em ICLR}, 2021.

\bibitem[\protect\citeauthoryear{Joshi \bgroup \em et al.\egroup }{2022}]{gcrd}
Chaitanya~K Joshi, Fayao Liu, Xu~Xun, Jie Lin, and Chuan~Sheng Foo.
\newblock On representation knowledge distillation for graph neural networks.
\newblock {\em IEEE Transactions on Neural Networks and Learning Systems},
  2022.

\bibitem[\protect\citeauthoryear{Kipf and Welling}{2017}]{gcn}
Thomas~N. Kipf and Max Welling.
\newblock Semi-supervised classification with graph convolutional networks.
\newblock In {\em ICLR}, 2017.

\bibitem[\protect\citeauthoryear{Lesort \bgroup \em et al.\egroup
  }{2020}]{kd_robotics}
Timothee Lesort, Vincenzo Lomonaco, Andrei Stoian, Davide Maltoni, David
  Filliat, and Natalia Diaz-Rodriguez.
\newblock Continual learning for robotics: Definition, framework, learning
  strategies, opportunities and challenges.
\newblock {\em Information fusion}, 2020.

\bibitem[\protect\citeauthoryear{Li \bgroup \em et al.\egroup
  }{2019}]{deepgcns}
Guohao Li, Matthias Muller, Ali Thabet, and Bernard Ghanem.
\newblock Deepgcns: Can gcns go as deep as cnns?
\newblock In {\em ICCV}, 2019.

\bibitem[\protect\citeauthoryear{Li \bgroup \em et al.\egroup
  }{2022a}]{text_classification_1}
Quan Li, Xiaoting Li, Lingwei Chen, and Dinghao Wu.
\newblock Distilling knowledge on text graph for social media attribute
  inference.
\newblock In {\em SIGIR}, 2022.

\bibitem[\protect\citeauthoryear{Li \bgroup \em et al.\egroup
  }{2022b}]{egnn_explanation}
Yuan Li, Li~Liu, Guoyin Wang, Yong Du, and Penggang Chen.
\newblock Egnn: Constructing explainable graph neural networks via knowledge
  distillation.
\newblock {\em Knowledge-Based Systems}, 2022.

\bibitem[\protect\citeauthoryear{Liu \bgroup \em et al.\egroup }{2019}]{kd_cv}
Yifan Liu, Ke~Chen, Chris Liu, Zengchang Qin, Zhenbo Luo, and Jingdong Wang.
\newblock Structured knowledge distillation for semantic segmentation.
\newblock In {\em CVPR}, 2019.

\bibitem[\protect\citeauthoryear{Ma \bgroup \em et al.\egroup
  }{2022}]{ma2022hierarchical}
Yihong Ma, Patrick Gerard, Yijun Tian, Zhichun Guo, and Nitesh~V Chawla.
\newblock Hierarchical spatio-temporal graph neural networks for pandemic
  forecasting.
\newblock In {\em CIKM}, 2022.

\bibitem[\protect\citeauthoryear{Muller \bgroup \em et al.\egroup
  }{2019}]{label_smoothing}
Rafael Muller, Simon Kornblith, and Geoffrey~E Hinton.
\newblock When does label smoothing help?
\newblock In {\em NeurIPS}, 2019.

\bibitem[\protect\citeauthoryear{Phuong and Lampert}{2019}]{kd_theory_2}
Mary Phuong and Christoph Lampert.
\newblock Towards understanding knowledge distillation.
\newblock In {\em ICML}, 2019.

\bibitem[\protect\citeauthoryear{Qian \bgroup \em et al.\egroup
  }{2021}]{detect_illicit_drug_trafficker_1}
Yiyue Qian, Yiming Zhang, Yanfang Ye, Chuxu Zhang, et~al.
\newblock Distilling meta knowledge on heterogeneous graph for illicit drug
  trafficker detection on social media.
\newblock In {\em NeurIPS}, 2021.

\bibitem[\protect\citeauthoryear{Ren \bgroup \em et al.\egroup
  }{2021}]{social_event_detection}
Jiaqian Ren, Hao Peng, Lei Jiang, Jia Wu, Yongxin Tong, Lihong Wang, Xu~Bai,
  Bo~Wang, and Qiang Yang.
\newblock Transferring knowledge distillation for multilingual social event
  detection.
\newblock {\em arXiv preprint arXiv:2108.03084}, 2021.

\bibitem[\protect\citeauthoryear{Rezayi \bgroup \em et al.\egroup
  }{2021}]{edge}
Saed Rezayi, Handong Zhao, Sungchul Kim, Ryan~A Rossi, Nedim Lipka, and Sheng
  Li.
\newblock Edge: Enriching knowledge graph embeddings with external text.
\newblock In {\em NAACL}, 2021.

\bibitem[\protect\citeauthoryear{Ruiz \bgroup \em et al.\egroup
  }{2020}]{transferable_gnn}
Luana Ruiz, Luiz Chamon, and Alejandro Ribeiro.
\newblock Graphon neural networks and the transferability of graph neural
  networks.
\newblock In {\em NeurIPS}, 2020.

\bibitem[\protect\citeauthoryear{Stark \bgroup \em et al.\egroup
  }{2022}]{gnn_molecular_1}
Hannes Stark, Dominique Beaini, Gabriele Corso, Prudencio Tossou, Christian
  Dallago, Stephan Gunnemann, and Pietro Lio.
\newblock 3d infomax improves gnns for molecular property prediction.
\newblock In {\em ICML}, 2022.

\bibitem[\protect\citeauthoryear{Sun \bgroup \em et al.\egroup }{2019}]{kd_nlp}
Siqi Sun, Yu~Cheng, Zhe Gan, and Jingjing Liu.
\newblock Patient knowledge distillation for bert model compression.
\newblock In {\em EMNLP}, 2019.

\bibitem[\protect\citeauthoryear{Tian \bgroup \em et al.\egroup
  }{2022a}]{reciperec}
Yijun Tian, Chuxu Zhang, Zhichun Guo, Chao Huang, Ronald Metoyer, and Nitesh~V
  Chawla.
\newblock Reciperec: A heterogeneous graph learning model for recipe
  recommendation.
\newblock In {\em IJCAI}, 2022.

\bibitem[\protect\citeauthoryear{Tian \bgroup \em et al.\egroup
  }{2022b}]{recipe2vec}
Yijun Tian, Chuxu Zhang, Zhichun Guo, Yihong Ma, Ronald Metoyer, and Nitesh~V
  Chawla.
\newblock Recipe2vec: Multi-modal recipe representation learning with graph
  neural networks.
\newblock In {\em IJCAI}, 2022.

\bibitem[\protect\citeauthoryear{Tian \bgroup \em et al.\egroup
  }{2022c}]{recipe_rec_frontiers}
Yijun Tian, Chuxu Zhang, Ronald Metoyer, and Nitesh~V Chawla.
\newblock Recipe recommendation with hierarchical graph attention network.
\newblock {\em Frontiers in big Data}, 2022.

\bibitem[\protect\citeauthoryear{Tian \bgroup \em et al.\egroup
  }{2023a}]{hgmae}
Yijun Tian, Kaiwen Dong, Chunhui Zhang, Chuxu Zhang, and Nitesh~V Chawla.
\newblock Heterogeneous graph masked autoencoders.
\newblock In {\em AAAI}, 2023.

\bibitem[\protect\citeauthoryear{Tian \bgroup \em et al.\egroup
  }{2023b}]{nosmog}
Yijun Tian, Chuxu Zhang, Zhichun Guo, Xiangliang Zhang, and Nitesh~V Chawla.
\newblock Nosmog: Learning noise-robust and structure-aware mlps on graphs.
\newblock In {\em ICLR}, 2023.

\bibitem[\protect\citeauthoryear{Velickovic \bgroup \em et al.\egroup
  }{2018}]{gat}
Petar Velickovic, Guillem Cucurull, Arantxa Casanova, Adriana Romero, Pietro
  Li{\`{o}}, and Yoshua Bengio.
\newblock Graph attention networks.
\newblock In {\em ICLR}, 2018.

\bibitem[\protect\citeauthoryear{Wang and Yoon}{2021}]{kd_survey_2}
Lin Wang and Kuk-Jin Yoon.
\newblock Knowledge distillation and student-teacher learning for visual
  intelligence: A review and new outlooks.
\newblock {\em IEEE Transactions on Pattern Analysis and Machine Intelligence},
  2021.

\bibitem[\protect\citeauthoryear{Wang \bgroup \em et al.\egroup
  }{2017}]{kg_survey_2}
Quan Wang, Zhendong Mao, Bin Wang, and Li~Guo.
\newblock Knowledge graph embedding: A survey of approaches and applications.
\newblock {\em IEEE Transactions on Knowledge and Data Engineering}, 2017.

\bibitem[\protect\citeauthoryear{Wang \bgroup \em et al.\egroup }{2021}]{mulde}
Kai Wang, Yu~Liu, Qian Ma, and Quan~Z Sheng.
\newblock Mulde: Multi-teacher knowledge distillation for low-dimensional
  knowledge graph embeddings.
\newblock In {\em WWW}, 2021.

\bibitem[\protect\citeauthoryear{Wang \bgroup \em et al.\egroup }{2022}]{ckd}
Can Wang, Sheng Zhou, Kang Yu, Defang Chen, Bolang Li, Yan Feng, and Chun Chen.
\newblock Collaborative knowledge distillation for heterogeneous information
  network embedding.
\newblock In {\em WWW}, 2022.

\bibitem[\protect\citeauthoryear{Wu \bgroup \em et al.\egroup
  }{2020}]{gnn_survey_1}
Zonghan Wu, Shirui Pan, Fengwen Chen, Guodong Long, Chengqi Zhang, and S~Yu
  Philip.
\newblock A comprehensive survey on graph neural networks.
\newblock {\em IEEE transactions on neural networks and learning systems},
  2020.

\bibitem[\protect\citeauthoryear{Wu \bgroup \em et al.\egroup }{2022}]{kdga}
Lirong Wu, Haitao Lin, Yufei Huang, and Stan~Z Li.
\newblock Knowledge distillation improves graph structure augmentation for
  graph neural networks.
\newblock In {\em NeurIPS}, 2022.

\bibitem[\protect\citeauthoryear{Xu \bgroup \em et al.\egroup }{2019}]{gin}
Keyulu Xu, Weihua Hu, Jure Leskovec, and Stefanie Jegelka.
\newblock How powerful are graph neural networks?
\newblock In {\em ICLR}, 2019.

\bibitem[\protect\citeauthoryear{Yan \bgroup \em et al.\egroup
  }{2020}]{tinygnn}
Bencheng Yan, Chaokun Wang, Gaoyang Guo, and Yunkai Lou.
\newblock Tinygnn: Learning efficient graph neural networks.
\newblock In {\em KDD}, 2020.

\bibitem[\protect\citeauthoryear{Yang \bgroup \em et al.\egroup }{2020}]{lsp}
Yiding Yang, Jiayan Qiu, Mingli Song, Dacheng Tao, and Xinchao Wang.
\newblock Distilling knowledge from graph convolutional networks.
\newblock In {\em CVPR}, 2020.

\bibitem[\protect\citeauthoryear{Yang \bgroup \em et al.\egroup }{2021}]{cpf}
Cheng Yang, Jiawei Liu, and Chuan Shi.
\newblock Extract the knowledge of graph neural networks and go beyond it: An
  effective knowledge distillation framework.
\newblock In {\em WWW}, 2021.

\bibitem[\protect\citeauthoryear{Yang \bgroup \em et al.\egroup }{2022}]{gkd}
Chenxiao Yang, Qitian Wu, and Junchi Yan.
\newblock Geometric knowledge distillation: Topology compression for graph
  neural networks.
\newblock In {\em NeurIPS}, 2022.

\bibitem[\protect\citeauthoryear{Yu \bgroup \em et al.\egroup }{2022}]{sail}
Lu~Yu, Shichao Pei, Lizhong Ding, Jun Zhou, Longfei Li, Chuxu Zhang, and
  Xiangliang Zhang.
\newblock Sail: Self-augmented graph contrastive learning.
\newblock In {\em AAAI}, 2022.

\bibitem[\protect\citeauthoryear{Yun \bgroup \em et al.\egroup }{2022}]{lte4g}
Sukwon Yun, Kibum Kim, Kanghoon Yoon, and Chanyoung Park.
\newblock Lte4g: Long-tail experts for graph neural networks.
\newblock In {\em CIKM}, 2022.

\bibitem[\protect\citeauthoryear{Zhang \bgroup \em et al.\egroup
  }{2019}]{zhang2019your}
Linfeng Zhang, Jiebo Song, Anni Gao, Jingwei Chen, Chenglong Bao, and Kaisheng
  Ma.
\newblock Be your own teacher: Improve the performance of convolutional neural
  networks via self distillation.
\newblock In {\em CVPR}, 2019.

\bibitem[\protect\citeauthoryear{Zhang \bgroup \em et al.\egroup }{2020}]{rdd}
Wentao Zhang, Xupeng Miao, Yingxia Shao, Jiawei Jiang, Lei Chen, Olivier Ruas,
  and Bin Cui.
\newblock Reliable data distillation on graph convolutional network.
\newblock In {\em SIGMOD}, 2020.

\bibitem[\protect\citeauthoryear{Zhang \bgroup \em et al.\egroup }{2021}]{rod}
Wentao Zhang, Yuezihan Jiang, Yang Li, Zeang Sheng, Yu~Shen, Xupeng Miao, Liang
  Wang, Zhi Yang, and Bin Cui.
\newblock Rod: reception-aware online distillation for sparse graphs.
\newblock In {\em KDD}, 2021.

\bibitem[\protect\citeauthoryear{Zhang \bgroup \em et al.\egroup
  }{2022a}]{mskd}
Chunhai Zhang, Jie Liu, Kai Dang, and Wenzheng Zhang.
\newblock Multi-scale distillation from multiple graph neural networks.
\newblock In {\em AAAI}, 2022.

\bibitem[\protect\citeauthoryear{Zhang \bgroup \em et al.\egroup
  }{2022b}]{glnn}
Shichang Zhang, Yozen Liu, Yizhou Sun, and Neil Shah.
\newblock Graph-less neural networks: Teaching old mlps new tricks via
  distillation.
\newblock In {\em ICLR}, 2022.

\bibitem[\protect\citeauthoryear{Zheng \bgroup \em et al.\egroup
  }{2022}]{cold_brew}
Wenqing Zheng, Edward~W Huang, Nikhil Rao, Sumeet Katariya, Zhangyang Wang, and
  Karthik Subbian.
\newblock Cold brew: Distilling graph node representations with incomplete or
  missing neighborhoods.
\newblock In {\em ICLR}, 2022.

\bibitem[\protect\citeauthoryear{Zhou \bgroup \em et al.\egroup }{2021}]{hkd}
Sheng Zhou, Yucheng Wang, Defang Chen, Jiawei Chen, Xin Wang, Can Wang, and
  Jiajun Bu.
\newblock Distilling holistic knowledge with graph neural networks.
\newblock In {\em ICCV}, pages 10387--10396, 2021.

\bibitem[\protect\citeauthoryear{Zhuang \bgroup \em et al.\egroup
  }{2022}]{dfadgnn}
Yuanxin Zhuang, Lingjuan Lyu, Chuan Shi, Carl Yang, and Lichao Sun.
\newblock Data-free adversarial knowledge distillation for graph neural
  networks.
\newblock In {\em IJCAI}, 2022.

\end{thebibliography}
